\documentclass[conference]{IEEEtran}
\usepackage{amsmath,amssymb,amsfonts}
\usepackage{algorithmic}
\usepackage{graphicx}
\usepackage{textcomp}
\usepackage{xcolor}
\usepackage{booktabs}
\usepackage{amsmath}
\usepackage{graphicx}
\usepackage{caption}
\usepackage{multirow}
\usepackage{hyperref}

\def\BibTeX{{\rm B\kern-.05em{\sc i\kern-.025em b}\kern-.08em
    T\kern-.1667em\lower.7ex\hbox{E}\kern-.125emX}}
\begin{document}

\title{A Multimodal RGB and Events Dataset for Hand Detection in First-person View\\

}
 
\author{
\IEEEauthorblockN{Bharghav Kota}
\IEEEauthorblockA{\textit{Zurich University of Applied Sciences} \\
Wädenswil, Switzerland \\
kotv@zhaw.ch }
\and
\IEEEauthorblockN{Yulia Sandamirskaya}
\IEEEauthorblockA{\textit{Zurich University of Applied Sciences} \\
Wädenswil, Switzerland \\
sank@zhaw.ch}
}

\maketitle

\begin{abstract}
Existing hand detection algorithms work on images and the detection rate is restricted by the frame rate of the camera. In hand detection applications for moving robotic systems, conventional cameras cause motion blur, especially in  darker lightning conditions. We can leverage the use of event-based cameras which possess a high dynamic range, high temporal resolution, and low power consumption. Recent work has shown that using a stereo setup of an event-based and a frame-based cameras improves detection accuracy and the bandwidth-latency tradeoff. The main bottleneck in using event-based cameras in object detection and recognition tasks is a relatively low amount of training data. In this work, we propose a methodology and an exemplary synthetic event-based hand dataset from an egocentric, first-person view perspective. The data is synthesised from the existing RGB \textit{Egohands} dataset with v2e toolbox. Parameters of the v2e toolbox are varied to provide versions of the dataset with different lighting conditions and scales. Ground truth detections are generated with a finetuned YOLOv8 model which is applied to the RGB images in Egohands dataset and interpolated on the high-temporal resolution events. We use the  multimodal dataset to perform hand detections with the existing object detection algorithms which use a multimodal setup of event and RGB cameras and demonstrate performance comparable to the state-of-the-art.  \end{abstract}


\section{Introduction}
Unlike traditional cameras that output intensity frames, event cameras have emerged that output asynchronous pixel-by-pixel event streams \cite{gallego2020event}. These events are emitted when a photoreceptor circuit detects a change in light intensity at a particular pixel resulting in an event stream that is sparse and has a high temporal resolution \cite{tobi_dvs}. This makes event cameras to feature high dynamic range, microsecond latency and thus low motion blur, and low power consumption. 
A camera with these properties becomes indispensible in the field of robotics and embedded vision \cite{falanga2020dynamic,hagenaars2020evolved,sanket2020evdodgenet,sun2021autonomous},where the systems are power and computationally constrained. 

As an event camera does not generate images, traditional pretrained (e.g., convolutional) neural networks cannot be leveraged for performing detection or tracking tasks directly. There is a need to develop algorithms that take advantage of the sparsity and high temporal resolution of an event camera while competing with the performance metrics of a deep learning approach. The first step would be to rethink the data representation of an event stream and a wide variety of work has been done in this regard. CNNs could be leveraged by accumulating events in a particular time interval into frames \cite{gehrig2019end,rebecq2019high,tulyakov2021time,zhu2019unsupervised,slam}, but this approach reduces the advantage of an event camera's low latency \cite{slam}. A more robust yet another frame-based approach is a motion-compensated event frame \cite{stoffregen2019event} which involves warping the events to align with a chosen reference frame, based on a defined motion model. By doing so, the accurate spatial edge structures over extended time intervals can be preserved. Non frame-based approaches have also emerged that take advantage of the temporal resolution such as time surfaces \cite{lagorce2016hots}. This is a 2D representation where the most recent timestamp of an event that has occurred at each pixel is retained and normalized to a range [0,1]. Several non-learning based methods have taken advantage of time surfaces to do detection tasks and visual SLAM, e.g. event FAST \cite{mueggler2017fast}.

High-frequency hand detection and pose estimation are critical components in robotic systems particularly for tasks
involving intention recognition \cite{intention}, human-robot interaction and object handovers \cite{handover}. It is challenging to perform high rate-detections with RGB cameras because we are limited  by the frame-rate of the camera. Solutions 
to this would be to interpolate detections in the blind times between frames or
use high FPS cameras which would increase hardware costs,
power consumption and computation budget of a robot. Event-
based cameras can be used to perform high-rate detections
because of their ability to capture per-pixel intensity changes
at a microsecond level resolution and low latency advan-
tages \cite{gehrig2024low, Aegnn}. But event-based approaches currently face limitations
in accuracy due to two main factors: the sensors’ inability to
detect slowly changing signals and the inefficiency of existing
processing techniques that transform event streams into frame-
based formats for analysis using convolutional neural networks \cite{sun2022ess,perot2020learning,alonso2019ev}. 

\cite{Aegnn} proposed a novel method to process events sparsely and asynchronously as temporally evolving graphs. This model can be trained on batches of events, taking advantage of backpropagation, and allows hierarchical learning using standard graph neural networks algorithms. Unlike AEGNN that uses a monocular event stream to perform object recognition and detection, DAGr \cite{gehrig2024low} proposed a hybrid event and frame-based object detector that fused a CNN for frames with an asynchronous graph neural network for events \cite{gehrig2024low}. DAGr performs superior to AEGNN as it combines the advantages of frame-based processing for rich context information and high accuracy detection with the sparsity and high rate of an event stream, thus improving the latency-bandwidth tradeoff.    

DAGr and AEGNN have been tested on popular automotive event-based datasets like \textit{NCARS} \cite{hats} and \textit{DSEC} \cite{Gehrig21ral,Gehrig3dv2021}, which feature dynamic driving scenarios. Their strong performance in these settings suggests that similar approaches could be effective in robotics, especially in dynamic environments and on mobile robots. A key challenge in assistive, human-centered robotics is detecting human hands, which is crucial for tasks such as object handover and understanding user intent. However, there is currently a lack of event-based datasets that show hands in dynamic, first person view —situations that reflect how a robot would perceive and interact with the world around it. In this work, we propose a dataset \textit{EventEgoHands} that is derived from the frame based dataset - \textit{Egohands} \cite{b3}. The RGB dataset provides manually labeled ground truths for 2.12\% of all frames in video sequences. The proposed \textit{EventEgoHands} is multimodal with synchronised events and frames and we extend these ground truths to the entire dataset by fine-tuning a YOLOv8 model and running inference on the frames where the ground truths were not provided in \textit{EgoHands}. We then train \textit{EventEgoHands} on the DAGr model to perform high-rate hand detections.

\section{Background}

\subsection{Graph Neural Networks}

A graph is a data structure \( G = \{V, E\} \) consisting of nodes/vertices  \( V \)  and edges \( E \) that connect these nodes. Information can be stored on in the graph as \textit{node features} or \textit{edge features} \cite{gnnsurvey}.

\textbf{Message passing} with graphs involves exchanging information between nodes in a graph along the edges they're connected to. The purpose of message passing is to aggregate information from neighboring nodes to encode contextual graph information.

\textbf{Graph Convolution} \cite{gcn}: Through message passing, each node updates its representation by combining its own features with those of its connected nodes weighted by the graph structure. This enables graph-convolutional networks to capture both local and global dependencies.

\begin{equation}
H^{(l+1)} = \sigma\left( \tilde{D}^{- \frac{1}{2}} \tilde{A} \tilde{D}^{- \frac{1}{2}} H^{(l)} W^{(l)} \right)
\end{equation}

where:
\begin{itemize}
    \item $H^{(l)} \in \mathbb{R}^{N \times F_l}$ is the node feature matrix at layer $l$,
    \item $W^{(l)} \in \mathbb{R}^{F_l \times F_{l+1}}$ is the trainable weight matrix,
    \item $\tilde{A} = A + I$ is the adjacency matrix with added self-loops,
    \item $\tilde{D}$ is the diagonal degree matrix of $\tilde{A}$,
    \item $\sigma(\cdot)$ is an activation function such as ReLU.
\end{itemize}

This operation performs feature aggregation from a node's local neighborhood, normalized by the degrees of the nodes to prevent scale distortion and over-smoothing. It enables the model to learn representations that capture both node features and graph topology.

\subsection{Event Based Cameras}


 \begin{figure}[h]
 	\centering
 	\includegraphics[width=0.4\textwidth, keepaspectratio]{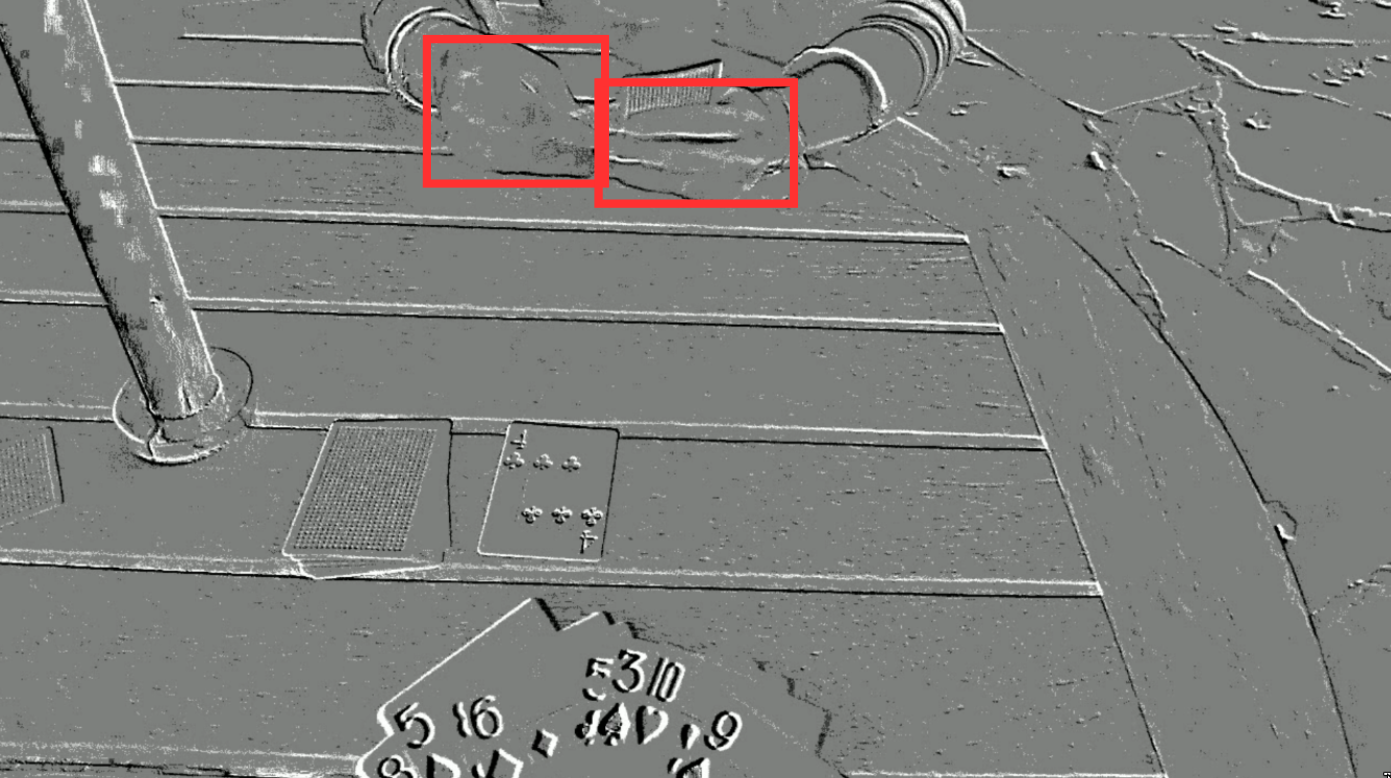} 	
    \caption[Event based cameras]{Accumulated event frame (event accumulation for 33ms) }
\label{fig:accumulated}
 \end{figure}

  \begin{figure}[h]
 	\centering
 	\includegraphics[width=0.4\textwidth, keepaspectratio]{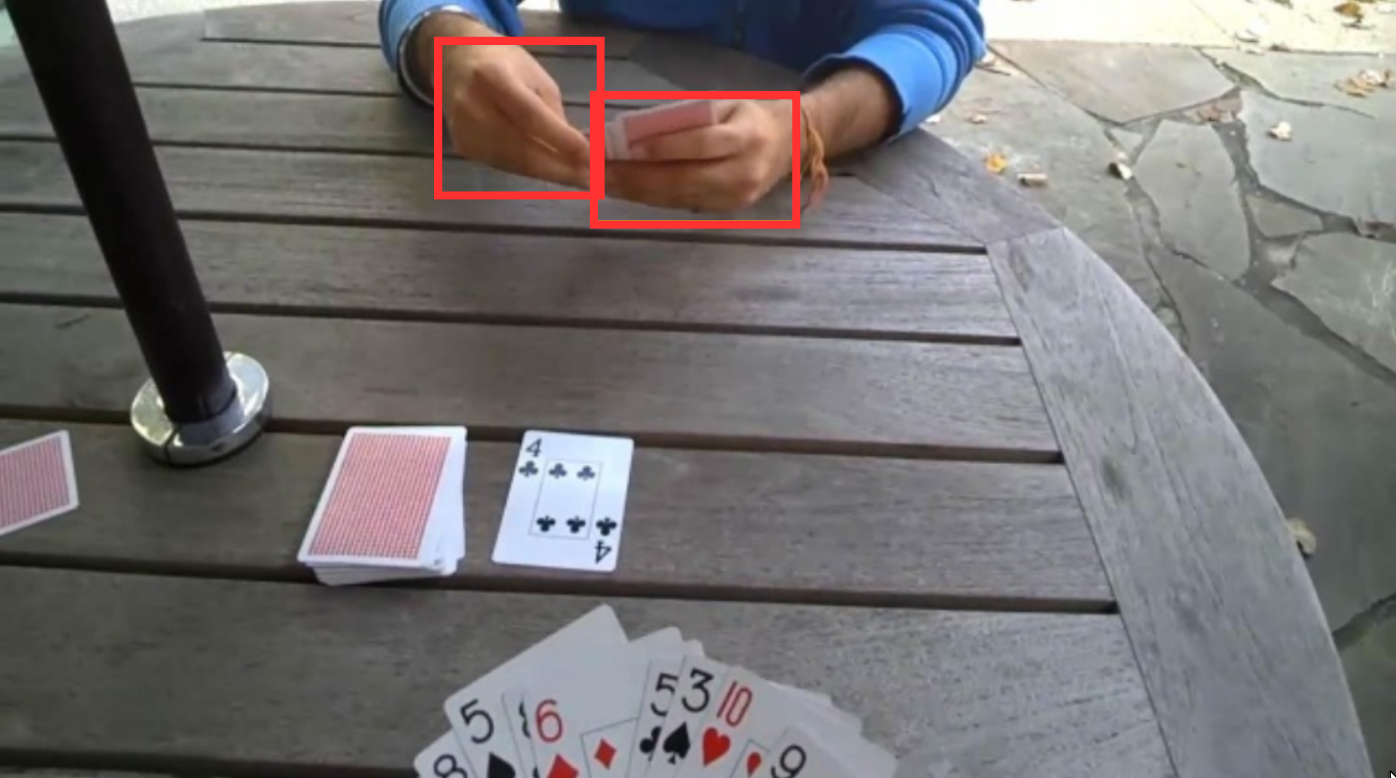} 	
    \caption[Event based cameras]{RGB frame from \textit{EgoHands}}
\label{fig:rgb_frame}
 \end{figure}

   \begin{figure}[h]
 	\centering
 	\includegraphics[width=0.4\textwidth, keepaspectratio]{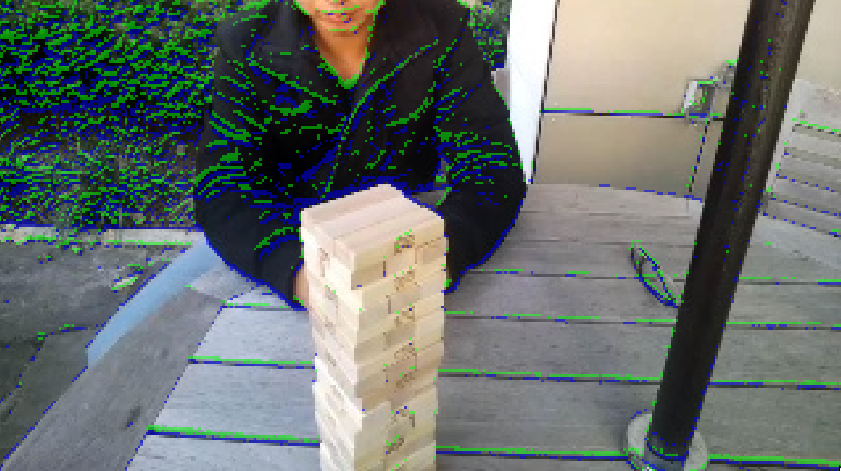} 	
    \caption[Event based cameras]{Events overlayed with RGB Frame}
\label{fig:overlay}
 \end{figure}

    \begin{figure}[h]
 	\centering
 	\includegraphics[width=0.4\textwidth, keepaspectratio]{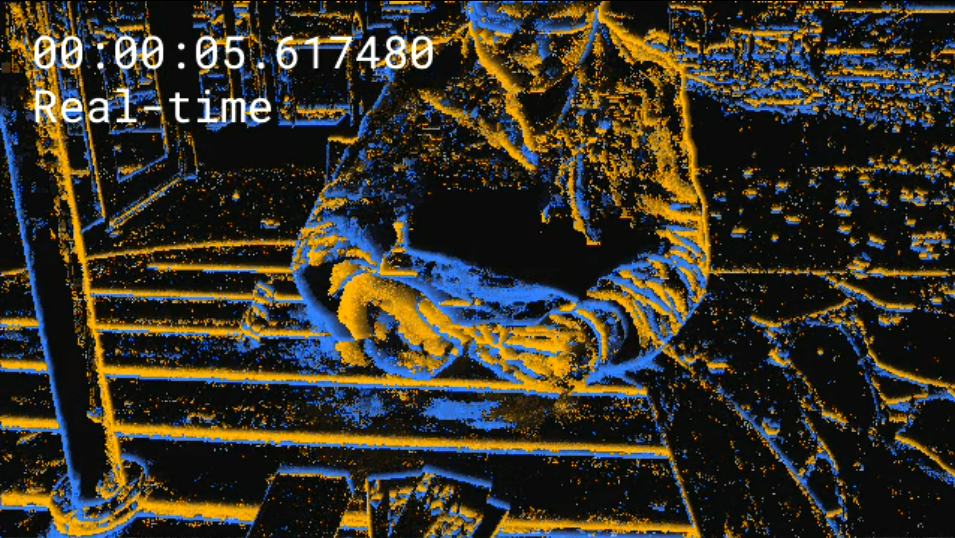} 	
    \caption[Event based cameras]{Faery render of \textit{Clean} events}
\label{fig:clean}
 \end{figure}

    \begin{figure}[h]
 	\centering
 	\includegraphics[width=0.4\textwidth, keepaspectratio]{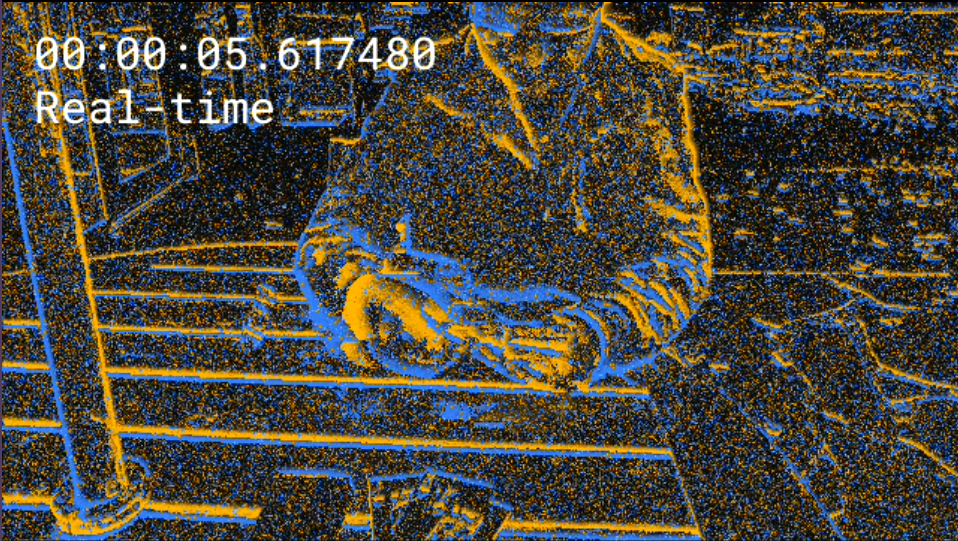} 	
    \caption[Event based cameras]{Faery render of \textit{Noisy} events}
\label{fig:noisy}
 \end{figure}

The Dynamic Vision Camera (DVS) pixel operates through a circuit sensitive to change in illumination which is governed by bias currents that define both the detection threshold and analog bandwidth \cite{tobi_dvs}. The incident light induces a logarithmic voltage $V_p$ change at the photoreceptor which is then inverted and amplified.  This voltage $V_d$ is then compared with the on and off thresholds and the pixel emits an event. This  triggers a reset that stores the updated intensity across the capacitor $C$. Understanding how the pixel circuit works is important to understand the impact of the parameter tuning in the v2e event-generation toolbox \cite{b2}.

\subsection{Event Graphs}

Event streams are typically represented with the address event representation where each event is in the format of $(x,y,t,p)$. $x$ and $y$ represent the pixel coordinates, $t$ is the timestamp of the event with microsecond resolution, and $p=\{0,1\}$ is the binary polarity to indicate a increase or decrease in light intensity. An event graph is constructed in a spatiotemporal space with $x, y, t$ as the axes where a node in the graph represents an event. The edges of the graph are constructed by joining nodes within a spatio-temporal distance $R$ from each other \cite{Aegnn}.

\subsection {v2e}
v2e (Video-to-Events) is a toolbox that generates realistic synthetic DVS events from intensity frames \cite{b2}. Unlike other event-generating simulators such as ESIM \cite{rebecq2018esim}  and adv2e \cite{jiang2024adv2e}, v2e incorporates serveral aspects of a real DVS behavior such as pixel-level Gaussian event threshold mismatch, intensity-dependant noise, finite intensity dependent bandwidth, temporal noise, and leak events \cite{leak}. As a result of simulating DVS non-idealities, v2e can better model pixels in bad lighting conditions which is an important application of the DVS. The pipeline of v2e receives intensity frames and optionally interpolates frames with the SuperSloMo model \cite{jiang2018super} which predicts the bidirectional optic flow vectors from consecutive frames to perform interpolation at any desired timestamp between the original inputs. It then computes logarithmic intensity of each pixel and detects changes in log intensity that exceed pixel-specific thresholds, trigerring synthetic ON or OFF events while also optionally adding temporal noise and simulating leak events.

\section{Dataset Pipeline}

The following subsections talk about the pipeline that was used to generate different versions of EventEgoHands dataset. \textit{EventEgoHands} is a synthetic dataset generated using the v2e (video-to-events) simulator from the existing RGB egocentric hands dataset: \textit{Egohands}.

\subsection{RGB Egohands Dataset description}\label{AA}

\begin{itemize}
    \item \textbf{Dataset Size}: The dataset consists of 48 videos, each 90 seconds long, with 2700 frames (30fps) and each frame is of the resolution 720x1280px.

    \item \textbf{Classes}: The dataset contains 4 class labels: y\textit{our left, your right, my left, my right.}
These labels reflect a distinction between ”your” and ”mine,” as the dataset was captured
with two individuals sitting opposite each other. The actions were recorded using a Google Glass device, and the labels are categorized from the perspective of the person whose
glasses were recording.

    \item \textbf{Labels}: From each of the 48 videos, 100 frames were randomly sampled, and the hands
were manually annotated. This process resulted in a substantial dataset with 15,053
ground-truth labeled hands.

\end{itemize}


\subsection{Generating Synthetic Events}

\begin{table}[htbp]
\centering
\caption{Overview of EventEgoHands and their versions}
\label{tab:egohands_versions}
\footnotesize 
\setlength{\tabcolsep}{5pt} 
\renewcommand{\arraystretch}{1.2} 
\begin{tabular}{@{}lccc@{}}
\toprule
\textbf{Version} & \textbf{Upsampling\_Factor} & \textbf{Scale\_Factor} & \textbf{Events Model} \\
\midrule
v1 & 1  & 1  & Clean \\
v2 & 1  & 4  & Clean \\
v3 & 1  & 2  & Clean \\
v4 & 41 & 2  & Clean \\
v5 & 41 & 2  & Noisy \\
v6 & 41 & 2  & Mixed \\
\bottomrule
\end{tabular}
\end{table}

\begin{table}[htbp]
\centering
\caption{v2e parameters for "clean" and "noisy" dataset versions. $\theta$ and $\sigma_\theta$ represent event threshold and threshold variation respectively.}

\label{tab:dataset_conditions}
\small
\renewcommand{\arraystretch}{1.2}
\begin{tabular}{@{}lcc@{}}
\toprule
\textbf{} & \textbf{Clean} & \textbf{Noisy} \\
\midrule
$\theta$            & (0.2$_\text{ON}$, 0.2$_\text{OFF}$)                            & (0.2$_\text{ON}$, 0.2$_\text{OFF}$) \\
$\sigma_\theta$     & 0.02 & 0.03 \\
Shot Noise          & 5 Hz                  & 0 Hz \\
Leak Events         & 0.1 Hz               & 0 Hz \\
Cutoff Freq.        & 0 Hz                  & 30 Hz \\
\bottomrule
\end{tabular}
\end{table}

EventEgoHands is generated by using the v2e toolbox and we provide various versions of this dataset with different event generation models of a DVS Pixel. Different versions of \textit{EventEgoHands} is summarised in \autoref{tab:egohands_versions}. \textit{Upsampling\_Factor} is a parameter that is calculated from the v2e parameter \textit{timestamp\_resolution} and controls temporal upsampling of the the video from the source fps to achieve at least the set timestamp resolution. The orginal resolution of the dataset (\textit{scale\_factor }1) is 720x1280px and we create different versions where the events and images are downsampled to 360x640px (scale\_factor 2) and 180x320px (scale\_factor 4).
``Clean" and ``Noisy" are parameter presets in the v2e toolbox where ``Clean'' turns off noise, sets unlimited bandwidth and makes threshold variation small, whereas ``Noisy" sets a limited bandwidth and adds leak events and shot noise. Leak events are ON events that real DVS pixels randomly emit caused by junction leakage and parasitic photocurrent in the change detector reset switch. Shot noise is the temporal noise rate of ON+OFF events in darkest parts of a scene which are reduced in the brighter parts. \autoref{fig:clean} and \autoref{fig:noisy} show frames rendered from \textit{Clean} and \textit{Noisy} event models respectively to give a comparison of the two event models. These renderings are generated at the exact same timestamp with the help of the \textit{Faery} library (\url{https://github.com/aestream/faery}). \autoref{tab:dataset_conditions} provides the values of the above mentioned parameters.

\section{Results}
\subsection{Dataset}

\begin{figure}[t]
 	\centering
 	\includegraphics[width=0.4\textwidth, keepaspectratio]{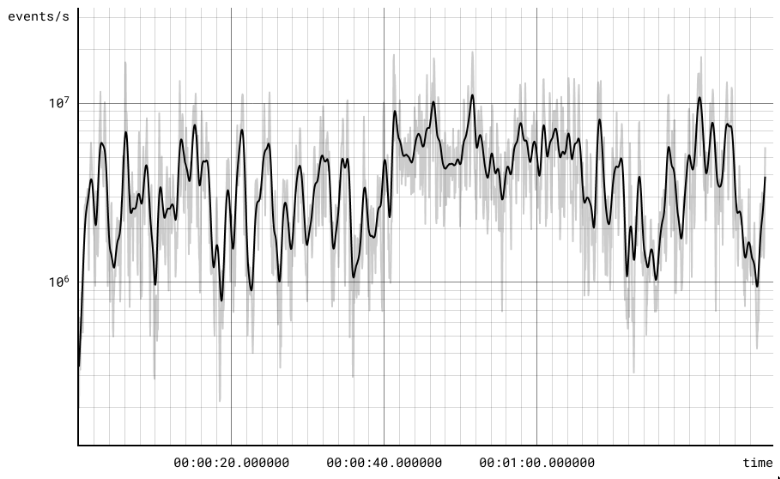} 	
    \caption[Event based cameras]{Event rate - Clean modelling }
\label{fig:plot_clean}
 \end{figure}

\begin{figure}[t]
 	\centering
 	\includegraphics[width=0.4\textwidth, keepaspectratio]{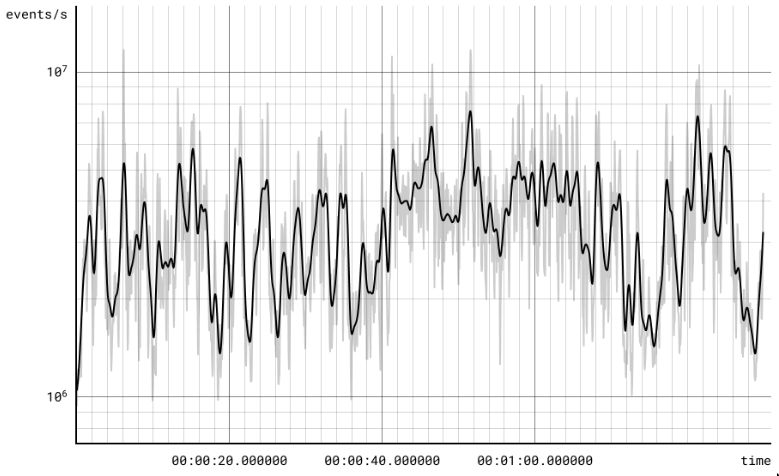} 	
    \caption[Event based cameras]{Event rate - Noisy modelling}
\label{fig:plot_noisy}
 \end{figure}

\begin{itemize}
    \item \textbf{Overview}: The dataset consists of 48 .h5 files for event streams which are synchronized with 48 RGB video sequences - each of 90secs, at 30fps. We also generate a key in the event file "\textit{timeidx}" which holds indices of the events that occur immediately after the timestamp of a particular frame. With the help of this indices list, we can speed up the process of slicing events exactly at the timestamp window to feed to the model for training. The frame resolution depends on the dataset version, which we provide at scales of 4, 2, and 1. For example, a scale of 4 indicates that the frame has been downsampled by a factor of 4. \autoref{fig:plot_clean} and \autoref{fig:plot_noisy} show the event rate plots (events/s over time) for the example sequence \textit{CARDS\_COURTYARD\_B\_T}, comparing the \textit{Clean} and \textit{Noisy} event models.

    \item \textbf{Classes}: Unlike \textit{EgoHands} we provide a single class ``hand" instead of differentiating between right and left hands.

    \item \textbf{Labels}:All hands in the dataset are labeled with bounding boxes at each frame timestamp, resulting in a total of 393,561 bounding boxes across 129,600 frames.

\end{itemize}
Refer to \autoref{tab:egohands_versions} for a comparision of different dataset versions with the respecting upsampling factor, scale factor and the DVS model used to generate events. \textit{EventEgoHands} dataset and training code is open-sourced at: \url{https://github.com/SynthSyntax/EventEgoHands}.

\subsection{Implementation details}
On EventEgoHands, we train the DAGr model with a batch size of 16, the learning rate
of $2 \times 10^{-4}$  AdamW optimizer \cite{adamw}. We train the network using a single image along with 33 ms of preceding events, aligning with the 30 Hz label frequency.

\subsection{EventEgoHands Hand Detection}

\begin{table}[htbp]
\centering
\caption{Comparison of different Egohands dataset versions on detection performance.}
\label{tab:mAP}
\small
\renewcommand{\arraystretch}{1.2} 
\begin{tabular}{@{}lccc@{}}
\toprule
\textbf{Version} & \textbf{mAP\textsubscript{50}} & \textbf{mAP\textsubscript{75}} & \textbf{mAP} \\
\midrule
v1 & 0.912 & 0.687 & 0.645 \\
v2 & 0.923 & 0.712 & 0.616 \\
v3 & 0.932 & 0.763 & 0.655 \\
v4 & 0.932 & 0.786 & 0.674 \\
v5 & 0.891 & 0.584 & 0.532 \\
v6 & 0.931 & 0.817 & 0.697 \\
\bottomrule
\end{tabular}
\end{table}

\begin{figure}[t]
 	\centering
 	\includegraphics[width=0.4\textwidth, keepaspectratio]{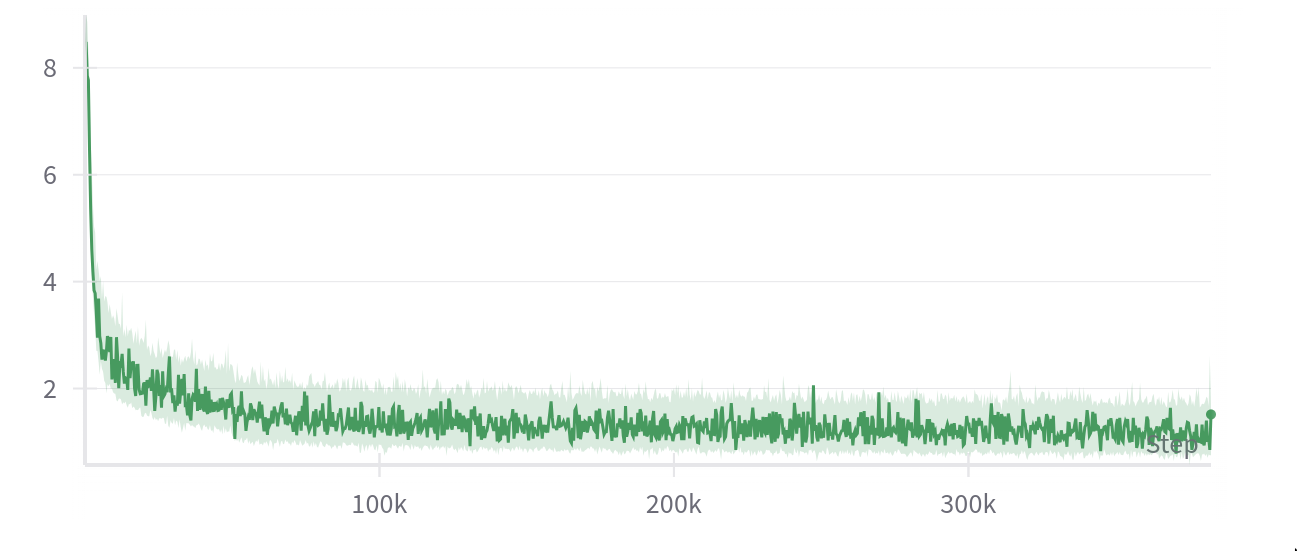} 	
    \caption[Event based cameras]{IoU loss convergence plot while training }
\label{fig:convergence}
 \end{figure}

\begin{figure}[t]
 	\centering
 	\includegraphics[width=0.4\textwidth, keepaspectratio]{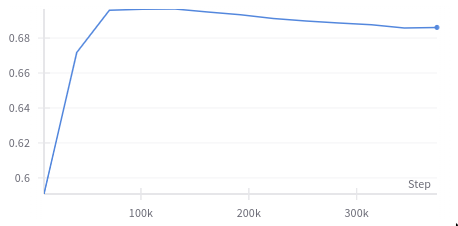} 	
    \caption[Event based cameras]{validation mAP metric plot}
\label{fig:validation}
 \end{figure}

This sections presents experiments done on the EventEgoHands dataset with the \textit{DAGr} model. We train this model from scratch with a single class "hand". The entire dataset consists of 48 videos and we create a test, train and validation split with 30, 10 and 8 videos respectively. We retain this split for all the dataset versions that are used to train the model. The weights obtained from different dataset versions are then used to run inference on a common test dataset and these hand detection metrics are presented in \autoref{tab:mAP}. This paper uses the COCO metrics \cite{coco} for testing the model where mAP (mean average precision) is calculated at various IoU (intersection over union) thresholds. mAP\textsubscript{50} is at an IoU of 0.5, mAP\textsubscript{75} as at an IoU of 0.75 and the overall mAP is averaged across IoU thresholds from 0.50 to 0.95 in steps of 0.05. \autoref{fig:convergence} shows the convergence plot of the IoU loss during training. The consistent decrease in loss indicates that the model is effectively learning to improve its bounding box predictions. These plots are derived in the case of using training data that is a mix of \textit{Clean} and\textit{ Noisy} events. Additionally, the validation mAP (.50:.05:.95) plot in \autoref{fig:validation} shows that the mAP peaks around 100k training steps, after which it begins to decline slightly. This suggests that the model starts to overfit beyond this point, as it continues to improve on the training data (as seen from the decreasing loss) but no longer generalizes as well to the validation set.

We trained the DAGr model on training data with ``Clean" and ``Noisy" presets in the v2e toolbox (refer to parameters in \autoref{tab:dataset_conditions}) and also used a training data which combined a mix of events in both conditions. It can be seen that the mAP obtained from v2 (scale\_factor of 4) has a lower mAP compared to all other versions with \textit{Clean} events. This result can be attributed to the downsampling being quite significant in events and the network fails to perform interframe detections effectively. We found scale2 to be ideal in retaining enough event data and at the same time compressing the original size of the dataset significantly thus increasing the training speed. Another interesting observation is the jump of accuracy between models trained with v3 and v4. This is because in v4, the events are generated by first upsampling the source fps of the video by a factor of 41 and then generating events synthetically. A more finegrained timestamp resolution of frames ensures that the events generated are modelled closer to the real DVS events and this gave an 0.02 mAP increment. Naturally, the mAP from Clean events in v4 is found to be higher compared to the model trained with \textit{Noisy} events (v5) because of ideal v2e parameters. Interestingly, training the model with data consisting of a combination of events from \textit{Noisy} and \textit{Clean} (v6) gives us another 0.02 mAP increment approximately. This result shows us that by incorporating a broad set of synthesis parameters, it improves model generalization on real event data and the difference in network performance between real and synthetic event data can be minimized.

\section{conclusion and further work}
This paper presents a multimodal events and frames dataset \textit{EventEgoHands} synthesised from the v2e toolbox. We provide various versions of this dataset by tuning v2e parameters and scales. Our experiments are run on Graph Neural Network based detection algorithm, trained with various dataset versions. We show that by simulating the noise and motion blur non-idealities of an event camera, we can bridge the gap between network performance in real and synthetic event data. Future work includes creating an event-based Hands dataset in first-person view with more variation in lighting conditions, skin tones and activity being performed. Further, this dataset can also be used to perform inference with evGNN \cite{b4}, an event-driven GNN based accelerator for vision on the edge for real-time robotic applications.

\bibliographystyle{IEEEtran}
\bibliography{references}

\end{document}